\documentclass[letterpaper]{article} 
\usepackage{aaai25}  
\usepackage{times}  
\usepackage{helvet}  
\usepackage{courier}  
\usepackage[hyphens]{url}  
\usepackage{graphicx} 
\usepackage{natbib}  
\usepackage{caption} 
\usepackage{algorithm}
\usepackage{algorithmic}

\usepackage{newfloat}
\usepackage{listings}
\DeclareCaptionStyle{ruled}{labelfont=normalfont,labelsep=colon,strut=off} 
\floatstyle{ruled}
\newfloat{listing}{tb}{lst}{}
\floatname{listing}{Listing}
\usepackage{rotating}
\usepackage{booktabs}    
\usepackage{siunitx} 
\usepackage{amsmath}

\usepackage[table]{xcolor}
\usepackage{multirow}

\usepackage{graphicx}  
\usepackage{subcaption}  

\usepackage{dirtytalk}

\title{When Explainability Meets Privacy:\\ An Investigation at the Intersection of Post-hoc Explainability and Differential Privacy in the Context of Natural Language Processing}
\author{
    Mahdi Dhaini, Stephen Meisenbacher, Ege Erdogan, Florian Matthes, Gjergji Kasneci
}
\affiliations{
    Technical University of Munich\\ School of Computation, Information and Technology\\ Department of Computer Science\\ Munich, Germany

    \{mahdi.dhaini, stephen.meisenbacher, ege.erdogan, matthes, gjergji.kasneci\}@tum.de
    
}

\usepackage{bibentry}

\begin{document}

\maketitle

\begin{abstract}
In the study of trustworthy Natural Language Processing (NLP), a number of important research fields have emerged, including that of \textit{explainability} and \textit{privacy}. While research interest in both explainable and privacy-preserving NLP has increased considerably in recent years, there remains a lack of investigation at the intersection of the two. This leaves a considerable gap in understanding of whether achieving \textit{both} explainability and privacy is possible, or whether the two are at odds with each other. In this work, we conduct an empirical investigation into the privacy-explainability trade-off in the context of NLP, guided by the popular overarching methods of \textit{Differential Privacy} (DP) and Post-hoc Explainability. Our findings include a view into the intricate relationship between privacy and explainability, which is formed by a number of factors, including the nature of the downstream task and choice of the text privatization and explainability method. In this, we highlight the potential for privacy and explainability to co-exist, and we summarize our findings in a collection of practical recommendations for future work at this important intersection.
\end{abstract}


\begin{links}
    \link{Code}{https://github.com/dmah10/xpnlp}
\end{links}

\section{Introduction}

Recent advances in Natural Language Processing (NLP) have seen bountiful and widespread improvements in the way natural language can be understood and generated. Such progress, hallmarked by the rapid developments enabled by Large Language Models (LLMs) and associated techniques, has powered novel applications in a variety of domains including education \citep{llm_education}, healthcare \citep{llm_healthcare}, and finance \citep{llm_finance}, as well as empowered non-technical users to explore the capabilities of Artificial Intelligence \citep{ai_non_technical}. These benefits, however, do not come for free, and various subfields of NLP currently work at the intersection of NLP and a number of human-centered topics, such as explainability \cite{danilevsky-xnlp-survey-2020}, privacy \citep{sousa2023keep}, bias \citep{nlp_bias}, fairness \citep{nlp_fairness}, and sustainability \cite{van2021sustainable}, among others.

Despite the recent democratization of LLM \textit{use}, a persistent challenge in the deployment of any language models relates to that of \textit{explainability}, which generally refers to the ability to interpret and communicate the decisions provided by a model. Explainability becomes paramount to the safe deployment of models, particularly in ensuring that model inputs can (reasonably) be traced or explained. Beyond this, explainability is not only a desired characteristic of a language model, but also a mandate (mainly for high-risk AI systems) under the recent EU AI Act regulation \citep{eu-ai-act}. One particularly useful candidate for fulfilling this mandate comes in the form of \textit{post-hoc explainability} \cite{madsen-xnlp-survey-2023, danilevsky-xnlp-survey-2020}, which comprises methods that serve to provide insights into traditionally \say{black-box} models. 

In a similar vein, the stark increase in LLM usage, particularly where users must interact with models hosted on external servers, i.e., in the cloud, has contributed to rising concerns of privacy \cite{9152761,wu2023unveiling,10198233,YAN2025100300}. Calls for privacy protection have been driven by increasingly strict data protection regulations (such as the GDPR); at the same time, privacy has been addressed in a plethora of research on privacy-preserving NLP \citep{yin-habernal-2022-privacy-nlp}, ranging from text privatization to private model training \cite{sousa2023keep}. Privacy-preserving techniques aim to mask both direct and indirect identifiers hidden within textual data, while also preserving utility for downstream tasks and applications \cite{mattern-etal-2022-limits, 10.1145/3485447.3512232}. One popular framework is Differential Privacy (DP) \cite{dwork2006differential}, which lends plausible deniability to text inputs by ensuring some level of indistinguishability between any two texts, usually achieved via the injection of random noise into text representations \cite{klymenko-etal-2022-differential}. This \say{noisification} can take place on many levels, such on the word level, or alternatively, via document-level rewriting.

Many recent works in Differentially Private Natural Language Processing (DP-NLP) focus on balancing the \textit{privacy-utility} trade-off as a key indicator for the effectiveness of a privatization method \cite{mattern-etal-2022-limits,utpala-etal-2023-locally}. Other works explored the trade-offs in other important aspects, such as text coherence \cite{10.1145/3485447.3512232} or user acceptability \cite{meisenbacher-etal-2025-investigating}. On the other hand, Explainable NLP (XNLP) has increasingly focused on the intersection of explainability with some aspects of trustworthy NLP, particularly fairness. This research has taken two primary directions: utilizing explainability as a tool for detecting bias in models \citep{fairnessxai-2024-towards, gallegos2024bias} and evaluating the fairness of explainability methods themselves \citep{dhaini-gender-disparity-2025}. However, specifically in the NLP domain, no works to the best of our knowledge consider the \textit{intersection of privacy and explainability} in terms of \textit{privacy-explainability trade-off}, namely, how the application of privatization methods affects the function of explainability methods. We argue that this consideration is a crucial one, particularly with the simultaneous legal mandate for both explainability and privacy.   

In this work, we are the first to investigate the interplay between privacy and explainability in the context of natural language. As a case study, we select two popular subfields: \textit{post-hoc feature attribution} methods and \textit{differentially private text rewriting}. In this, we explore the shift in explainability that can be observed when rewriting texts via DP, at various privacy levels and with fundamentally different rewriting mechanisms. We also consider the effect of downstream task specifics, such as model choice, model size, and fine-tuning task, particularly when varying the importance of privacy over explainability, and vice versa. To guide these experiments, we define one overarching research question:

\begin{quote}
    \textit{What is the impact of differentially private text rewriting on the post-hoc explainability of fine-tuned language models, and how can one quantify the privacy-explainability trade-off?}
\end{quote}

We learn that while a clear trade-off can typically be observed between privacy and explainability, there do exist configurations in which the two work synergistically. In this, we find that the factors of the downstream dataset and task, as well as the selected DP method and its privacy budget, are important in the quantification of the privacy-explainability trade-off. This leads us to create a collection of recommendations for both researchers and practitioners who wish to continue work at this important intersection. 

Concretely, our work makes the following contributions:
\begin{enumerate}
    \itemsep 0em
    \item We are the first to conduct an empirical investigation at the intersection of privacy and explainability, particularly in the context of natural language. We make our experimental code available for reproducibility.
    \item We provide insights into the complex interplay between post-hoc explainability and differentially private text rewriting, serving as a foundation for broader investigations at this intersection.
    \item We analyze our results to propose recommendations on best practices for important design choices, particularly when faced with the need for explainability \textit{and} privacy.
\end{enumerate}


\section{Background and Related Work}

\subsection{Post-hoc explainability methods}

Model-agnostic \textit{feature-attribution post-hoc} techniques have gained prominence due to their broad applicability \cite{jacovi-2023-trendsexplainableaixai}. These methodologies seek to determine the relative contribution of individual tokens to model predictions for specific inputs, employing either gradient-based approaches that leverage model derivatives with respect to inputs \citep{sundararajan2017axiomatic, simonyan2013deep}, or perturbation-based techniques \citep{ribeiro2016should, lundberg2017unified}.

The expanding significance of explainable NLP research is demonstrated through the expansion of comprehensive surveys addressing NLP explainability \citep{ wallace-xnlp-survey-2020, Zhao-xnlp-survey-2024, madsen-xnlp-survey-2023, zini-xnlp-survey-2020}. Moreover, given the deployment of NLP systems in critical applications including education \citep{Wu_survey_nlp_education_Li_Gao_Weng_Ding_2024}, healthcare \citep{johri-2025-clinical-llm-evaluation} and legal domains \citep{valvoda-towards-xai-legal-prediction-2024} where interpretability requirements are essential, specialized surveys have emerged focusing on explainability within particular NLP applications, such as fact verification \citep{kotonya-explainable-fact-checking-survey-2020}, and specific methodological approaches in NLP explainability \citep{mosca-shap-survey-2022}. These comprehensive reviews underscore the extensive utilization of post-hoc methodologies across NLP applications.

Additionally, feature-attribution post-hoc explanation techniques serve as primary methods within explainability platforms and computational frameworks documented in recent literature \citep{cleverxai-2022, Li-m4-xai-benchmark-2024, attanasio-etal-2023-ferret, inseq2023}. These comprehensive frameworks characteristically integrate multiple post-hoc explanation algorithms while accommodating various data modalities Machine Learning (ML) architectures, including pre-trained Language Models (PLMs).

\subsection{Differential Privacy in NLP}
The notion of \textit{Differential Privacy} was first proposed in the context of relational databases \cite{dwork2006differential}, where the primary goal was to protect the participation of an \textit{individual} in the dataset. More specifically, privacy preservation occurs in the sense that information about such an individual cannot be accurately inferred within some bound. This is formalized via the following inequality, for any databases $D_1$ and $D_2$ differing in exactly one element, any $\varepsilon > 0$, any computation or function $\mathcal{M}$, and all $\mathcal{S} \subseteq Range(\mathcal{M})$:
\(\frac{Pr[\mathcal{M}(D_1) \in \mathcal{S}]}{Pr[\mathcal{M}(D_2) \in \mathcal{S}]} \le e^{\varepsilon}\).
Intuitively, DP ensures that there exists some level of \textit{indistinguishability} between any two neighboring databases (differing in one element), thus protecting the individual. This \textit{privacy level} is governed by the $\varepsilon$ parameter, also known as the \textit{privacy budget}.

This form of DP is known as $\varepsilon$-DP, and the notion above refers to \textit{global} DP. Another notion, which we focus on in this work, is that of \textit{local} DP (LDP) \cite{4690986}. In the local setting, we assume that the central curator, i.e., the one who is to possess the complete dataset, is not trusted. As a solution, DP is ensured at the user level; however, since the entirety of the dataset is not yet known, LDP imposes a much stricter indistinguishability requirement, i.e., between \textit{any} potential neighbor. This differs from the global notion, since neighboring databases only refer to those resulting from the dataset $D$. Formally, for finite space $\mathcal{P}$ and $\mathcal{V}$, and for all $x, x' \in \mathcal{P}$ and all $y \in \mathcal{V}$:
\(\frac{Pr[\mathcal{M}(x) = y]}{Pr[\mathcal{M}(x') = y]} \le e^\varepsilon\)
Hence, an observed output cannot be attributed to a specific input with a high probability. While this notion is clearly stricter, it allows for a quantification of a privacy guarantee on the local, single datapoint level without the need for an aggregated dataset.

The translation of DP into the realm of NLP initially brought about numerous research challenges \cite{klymenko-etal-2022-differential}, chief among them the reasoning of who the \textit{individual} is when considering textual data, and how to quantify neighboring \say{datasets}. Despite these challenges, a great deal of recent works have proposed innovative methods for the integration of DP into the NLP pipeline \cite{hu-etal-2024-differentially}, ranging for text anonymization and obfuscation to DP training of language models. Particularly considering \textit{text privatization}, many recent methods interpret the task from the \textit{rewriting} perspective, where a sensitive input text is rewritten under DP guarantees to produce a private output text. Such methods operate at various lexical levels, including the word level \cite{10.1145/3336191.3371856,carvalho2023tem}, the token-level during language model generation \cite{utpala-etal-2023-locally,meisenbacher-etal-2024-dp}, and on full texts \cite{igamberdiev-habernal-2023-dp}. Considering the differences in how these mechanisms operate, it becomes important to consider the nature of the DP guarantee when performing comparative analyses \cite{vu-etal-2024-granularity}.

Recent works in DP-NLP highlight persistent challenges, such as the generation of coherent and correct outputs \cite{mattern-etal-2022-limits}, ensuring comparability \cite{igamberdiev-etal-2022-dp}, and quantifying the benefit DP brings over non-DP methods \cite{meisenbacher-matthes-2024-thinking}. However, to the best of our knowledge, no existing works question the effect of DP methods on the \textit{explainability} of texts, or more specifically, of models trained on privatized texts. We see this to be a considerable gap, particularly for DP text privatization in important domains, such as the medical domain, where explainability is also important in conjunction with privacy. 

\subsection{Privacy meets Explainability} 

\paragraph{Privacy in Explainable ML.} 
The privacy implications of explainable ML have received attention in recent years but remain underexplored. Most of the work on the intersection of explainability and privacy in ML focuses on the inherent privacy risks in explanations. These works include studies that investigate how model explanations reveal sensitive information from the training data and how this can be exploited using membership inference attacks. For instance, \citet{shokri_privacy_explanations_2021} demonstrated that variance in backpropagation-based explanations can reveal whether a data point was used during training, exposing membership information. \citet{Duddu_xai_dp_attack_2022} extended these concerns by inferring sensitive attributes like gender and race directly from model explanations, while \citet{Liu_tell_me_more_2024} shows how explanations can amplify membership inference risks by exploiting differences in model robustness under attribution-guided perturbations. Similarly, \citet{Luo_xai_dp_attack_2022} showed that private input features can be reconstructed using Shapley-value explanations. Other works include highlighting how explanations can enhance model inversion attacks when used as auxiliary inputs \citep{Zhao_xai_dp_attack_2021}, and introducing a membership inference attack for counterfactuals relying on distances between original inputs and their counterfactual counterparts \citep{Pawelczyk_xai_dp_attack_2023}. 

\citet{shokri_privacy_explanations_2021} investigated the privacy risks of gradient-based explanation methods—Gradient \citep{gradient} and Integrated Gradients \citep{sundararajan2017axiomatic}—across four tabular and two image datasets. They showed that these methods can leak information about training examples, increasing vulnerability to membership inference attacks that use full feature attributions as inputs to an attack model. In contrast, perturbation-based methods such as LIME and SmoothGrad \citep{smoothGrad} were found to be more resistant, likely due to their reliance on input perturbations rather than gradients, which may capture subtle distinctions between training and non-training points. \citet{Liu_tell_me_more_2024} further examined membership inference risks for multiple post-hoc explainers in the context of image data.
 
\paragraph{Defenses and countermeasures.} 
to protect again inference attacks, some works in the literature used approaches that alter the training process such as DP-SGD \cite{Liu_tell_me_more_2024} or approaches that perturb confidence scores of the target models output for each input by adding noise and then convert the perturbed confidence scores into adversarial examples for the attack models, such as MemGuard \cite{Jia_memguard_defense_2019}. However such approaches either prove to be ineffective against inference attacks (such as MemGuard) \citep{Liu_tell_me_more_2024} or decrease the inference attack performance in the expense of severe degradation of the model utility as well as explanation quality such as DP-SGD and thus presenting a large trade-off between defense capability and  utility and explainability performance \cite{Liu_tell_me_more_2024}. 


\paragraph{Privacy in Explainable NLP.}
Prior work has primarily examined the privacy risks of model explanations in the context of tabular and image-based datasets. However, the intersection of privacy and explainability in NLP remains largely unexplored. In contrast to these studies, our work focuses specifically on textual datasets and NLP applications. Rather than analyzing privacy leakage from explanations, we empirically investigate whether data-level DP can be applied to achieve reasonable privacy guarantees while maintaining acceptable trade-offs in model utility and explanation quality. To the best of our knowledge, this is the first study to systematically evaluate the trade-off between data-level DP and the quality of post-hoc feature attribution explanations (in terms of their faithfulness) in NLP models.



\section{Experimental Setup}


\subsection{Datasets}
To guide our experiments, we select three datasets, which vary in size and domain. These datasets are introduced in the following, alongside their associated downstream tasks.

\paragraph{SST-2.}
The Stanford Sentiment Treebank dataset (SST-2) \citep{socher-etal-2013-recursive} comprises short texts originating from movie reviews, and it was popularized due to its inclusion in the \textsc{GLUE} benchmark \cite{wang-etal-2018-glue}. Each text is labeled according to its \textit{sentiment}, i.e., either positive or negative, creating a two-class binary classification task. We use the complete training split of the dataset, composed of 67,349 records, with an average word length of 9.41.

\paragraph{AG News.}
The AG News corpus contains over one million news articles from the over 2000 news sources. We utilize the subset as prepared by \citet{zhang2015character}, which contains 120k news articles from four news domains: \textit{world}, \textit{sports}, \textit{business}, and \textit{sci/tech}. We take a 50\% random sample (seed=42) for a final dataset of 60000 news articles, with an average word length of 43.90.

\paragraph{Trustpilot Reviews.}
The \textit{Trustpilot} corpus is a large-scale collection of user reviews. The corpus prepared by \citet{10.1145/2736277.2741141} tags each review with the stars provided (1-5), which we simplify to negative reviews (1-2 stars) and positive reviews (5 stars). We specifically only take reviews from the US split of the dataset (\textit{en-US}), and use a 10\% random sample. This results in a dataset of 29,490 reviews, with an average word length of 59.75.

\subsection{DP Methods} \label{sec:dp_methods}
We select three local DP text rewriting methods for our experiments, which are introduced in the following.

\paragraph{\textsc{TEM}.}
\textsc{TEM} \cite{carvalho2023tem} is a word-level DP mechanism which leverages a generalized notion called \textit{metric} DP. This generalization is useful for metric spaces, such as with word embeddings, wherein the indistinguishability requirement between words is scaled by their distance in the space. \textsc{TEM} improves upon previous approaches by employing a \textit{truncated exponential mechanism}, allowing for higher utility word replacements. To privatize a complete text, each component word is privatized one-by-one. Following the original work, we choose the privacy budgets of $\varepsilon \in \{1, 2, 3\}$, which refer to the budgets \textit{per word}.

\paragraph{DP-Prompt.}
Leveraging the generative capabilities of LLMs, \citet{utpala-etal-2023-locally} introduce \textsc{DP-Prompt}, a method for producing private outputs texts with local DP guarantees by modeling privatization as a \textit{paraphrasing} task. Internally, the \textsc{DP-Prompt} mechanism operates by applying DP at each token generation, specifically by using the temperature parameter as an equivalent DP mechanism to the Exponential Mechanism \cite{4389483}. Following the original work, we test three temperature values, $T \in \{1.75, 1.5, 1.25\}$. This translates to the per-token $\varepsilon$ values of $\varepsilon \in \{118, 137, 165\}$, using the implementation provided by \citet{meisenbacher-etal-2024-dp}, which employs a \textsc{flan-t5-base} model \cite{chung} as the underlying privatization model.

\paragraph{DP-BART.}
\textsc{DP-BART} is a document-level LDP text rewriting mechanism proposed by \citet{igamberdiev-habernal-2023-dp}. It leverages a pretrained \textsc{BART} model \cite{lewis-etal-2020-bart}, applying calibrated Gaussian noise in the latent representation space, where the decoder then decodes the noisy encoded vector to generate a private text. In doing so, \textsc{DP-BART} rewrites texts with a single document guarantee, albeit usually requiring higher privacy budgets for meaningful output. As such we choose $\varepsilon \in \{500, 1000, 1500\}$, and we use the original \textsc{DP-BART-CLV} variant.

\paragraph{Privatization Procedure.}
For \textsc{DP-BART}, we use the mechanism on the primary text column of each of our chosen datasets. This is repeated for each of the three privacy budgets, creating three private counterparts to the original data. Using \textsc{DP-Prompt} likewise produces full texts as a result of the generative process. Finally, \textsc{TEM} is run sequentially on all component words of an input text, which are tokenized using the \textsc{nltk.word\_tokenize} function. The list of outputs from the mechanism are then reconstructed into a single string via simple concatenation. In total, we thus produce nine private variants of the original datasets, for a total of 30 datasets (3 original baselines + 27 private counterparts).

\paragraph{A note on privatization and comparability.}
We caution that in the selection of privacy budgets for each DP mechanism, and for the resulting datasets based on these decisions, we do not ensure any comparability between the three selected methods. Due to the different manners in which DP is ensured across these three methods, as well as the intricacies involved with comparing different DP notions (e.g., MDP vs DP), we out-scope such comparisons. Instead, we focus on analyzing the downstream effects on DP rewriting \textit{within} each method (i.e., across privacy budgets).

We also note that the choice of privacy budgets ($\varepsilon$ values) are motivated by the choices taken by the authors of the original works. We do not work to normalize these values, i.e., by normalizing the logit values used in \textsc{DP-Prompt}. While this would be a useful step in reporting more usable and reasonable privacy budgets, we endeavor to test the DP methods as presented originally. As such, we do not report on or analyze the relative strength of underlying DP guarantee, as this is an active area of DP-NLP research.

\subsection{Models}
We utilize a number of pretrained encoder-only language models, which varying in architecture and model size. In this, we empirically measure the effect of fine-tuning on DP rewritten datasets, namely, the downstream effect of this process on the explainability of these models (measured by our chosen metrics). Table~\ref{tab:models_details} 
presents the information and details of the pretrained models used in our experiments to fine-tune on our three chosen classification datasets.

\begin{table}[h!]
    \scriptsize
    \centering
    \begin{tabular}{lll}
        \toprule
         \textbf{Name} & \textbf{Type} & \textbf{\# of Parameters} \\
         \midrule
        \textsc{BERT-base-cased} & Encoder-only & $\sim$110M \\
        \textsc{BERT-large-cased} & Encoder-only & $\sim$335M \\
         \textsc{RoBERTa-base} & Encoder-only & $\sim$125M  \\
         \textsc{RoBERTa-large} & Encoder-only & $\sim$355M \\
         \textsc{DeBERTa-base} & Encoder-only & $\sim$139M \\
         \bottomrule
    \end{tabular}
    \caption{\label{tab:models_details}Details of the experimental models.}
\end{table}

\subsection{Explainability Methods}

We include four post-hoc feature attribution methods in our experiments: \textbf{Gradient}
involves computing the gradient of the output with respect to the input features; \textbf{Integrated Gradient} instead integrates the gradients over a path from a baseline input to the explained input; \textbf{SHAP} \citep{lundberg2017unified} approximates the Shapley value of each feature, a concept from cooperative game theory that measures the ``contribution" of each feature by considering different coalitions of features and how much each feature contributes to the outcome;  \textbf{LIME} \citep{ribeiro2016should} attempts to replicate the model's behavior locally around the explained input with a linear model that is easy to explain. 

\subsection{Evaluating Explanations}

We evaluate the post-hoc explanations using two different methods each of measuring the \textbf{comprehensiveness} and \textbf{sufficiency} of explanations. Both metrics effectively attempt to quantify the \textit{faithfulness} of explanations, i.e. how accurately they reflect the underlying decision process of a model \citep{jacovi-goldberg-2020-faithfulness}. Faithfulness is one of the most important desideratum for explanations \citep{danilevsky-xnlp-survey-2020, lyu-etal-2024-faithful-survey-nlp} as high faithfulness indicates that the explanation accurately reflects the model’s decision-making process for a given prediction.

In their simplest forms, comprehensiveness metrics measure the change in output probabilities for the true class when the top-$k$ tokens with respect to the feature attribution scores are removed from the input, and sufficiency metrics measure the change when only the top-$k$ tokens are given as input to the model. We then compute the AOPC by averaging the values by $k$ is varied. We denote these metrics \textbf{AOPC-Comprehensiveness} and \textbf{AOPC-Sufficiency}.

Hard-removing the tokens might lead to out-of-distribution inputs for the model, which could adversely affect model performance during the evaluation of the explanations \citep{aopc_unfaithfulness_2022, aopc_unfaithfulness_2022_a}. To address this potential problem, we also include \textbf{Soft Sufficiency} and \textbf{Soft Comprehensiveness} metrics \citep{aopc-limitations-2023} in our evaluations. For the soft versions of these metrics, rather than removing a token entirely, a fraction of each token's embeddings is masked based on that token's importance score, i.e. the masked token is computed as $\mathbf{x}' = \mathbf{x} \odot \mathbf{e}$ where $\mathbf{e}_i \sim \text{Bernoulli}(q)$ with $q$ being the importance score if the token is kept (sufficiency) and one minus the score if it is removed (comprehensiveness).

\subsection{Composite Score}

Understanding that the trade-off between privacy and explainability is not always equally weighted, we design a metric that allows one to \textit{shift} the importance of either utility or explainability, while still viewing both in the same light. We compute a \textit{composite score} for each model $m$ and weight $\alpha\in\{0.25,0.5,0.75\}$ as
$\mathrm{CS}(m,\alpha)=\alpha\,\widehat{\mathrm{F1}}(m)+(1-\alpha)\,\widehat{E}(m)$,
where
$\widehat{E}(m)=\tfrac{1}{4}\bigl(\widehat{C}_{s}(m)+(1-\widehat{S}_{s}(m))+\widehat{C}_{a}(m)+(1-\widehat{S}_{a}(m))\bigr)$,
and $\widehat{x}$ denotes min–max normalization of $x$, subscripts $s$ and $a$ indicate “soft” vs.\ “AOPC,” $C$ is comprehensiveness, and $S$ is sufficiency (inverted via $1-\widehat{S}$ so that larger values are always better).\\
{\scriptsize
\begin{displaymath}
    \text{CS}(m, \alpha) = \alpha \hat{F1} + (1-\alpha)\hat{E}\\
\quad \hat{E} = \tfrac14\big(\hat{C}_s + (1-\hat{S}_s) + \hat{C}_a + (1-\hat{S}_a)\big)
\end{displaymath}
}

The composite score ranges from 0 to 1, where higher values (closer to 1) indicate near 1 means better overall performance (based on the selected $\alpha$ value). We intentionally define the score in this normalized form to enable a straightforward and consistent comparison across our experiments. Additionally, with this score, one can vary $\alpha$ based on the relative weight between utility and explainability. For example, an $\alpha$ of 0.25 would imply that the explainability faithfulness metrics are considerably more important, as opposed to a value of 0.75 where the utility score (F1) takes precedence.

\section{Results and Analysis}

Our results are presented across several tables and figures. Tables~\ref{tab:scores_alpha_25_avg}, \ref{tab:scores_alpha_50}, and \ref{tab:scores_alpha_75} report composite scores (mean$_{\scriptstyle\text{std}}$) for $\alpha = 0.25$, $0.5$, and $0.75$, respectively. Scores are averaged over five PLMs (\textsc{BERT-base}, \textsc{BERT-large}, \textsc{RoBERTa-base}, \textsc{RoBERTa-large}, and \textsc{DeBERTa-base}), and the \textit{Avg} row represents the mean and pooled standard deviation across the four explainers (Gradient, IG, LIME, and SHAP) for each (dataset, DP-$\varepsilon$) pair. In each row, the highest composite score is underlined. Columns are grouped by DP method (None, DPB, DPP, and TEM) and color-coded for comparison: grey for no DP, cyan for DPB, orange for DPP, and green for TEM. Color intensity reflects score magnitude, with darker shades indicating higher values. For instance, in Table~\ref{tab:scores_alpha_25_avg}, the most saturated cell in the TEM columns highlights the top-scoring explainer–DP-$\varepsilon$ combination (e.g., TEM3 with SHAP) for a given dataset and $\alpha$ value (e.g., AG News, $\alpha = 0.25$). This distinction in coloring is used to reflect the fact that DP methods are not directly comparable due to differences in mechanisms and assumptions (see Section~\ref{sec:dp_methods}). These tables support analysis of trends across explainers and privacy configurations.

\begin{table*}[ht!]
\small
\scriptsize
\setlength{\tabcolsep}{3pt}
\centering
\resizebox{0.7\linewidth}{!}{
\begin{tabular}{l l c c c c c c c c c c}
\toprule
Dataset & Expl & None & DPB500 & DPB1000 & DPB1500 & DPP118 & DPP137 & DPP165 & TEM1 & TEM2 & TEM3 \\
\midrule
\multirow{5}{*}{AG News} & G& 
  \cellcolor{gray!79}$0.705_{0.19}$ & 
  \cellcolor{cyan!14}$0.286_{0.14}$ & 
  \cellcolor{cyan!81}$0.655_{0.01}$ & 
  \cellcolor{cyan!90}$0.705_{0.01}$ & 
  \cellcolor{orange!82}$0.758_{0.01}$ & 
  \cellcolor{orange!85}$0.771_{0.02}$ & 
  \cellcolor{orange!85}$0.769_{0.02}$ & 
  \cellcolor{green!17}$0.333_{0.22}$ & 
  \cellcolor{green!57}$0.598_{0.17}$ & 
  \cellcolor{green!84}\underline{$0.777_{0.03}$} \\
 & IG&   
  \cellcolor{gray!51}$0.600_{0.13}$ & 
  \cellcolor{cyan!7}$0.245_{0.11}$ & 
  \cellcolor{cyan!57}$0.526_{0.02}$ & 
  \cellcolor{cyan!67}$0.577_{0.01}$ & 
  \cellcolor{orange!50}$0.610_{0.01}$ & 
  \cellcolor{orange!53}$0.620_{0.02}$ & 
  \cellcolor{orange!52}$0.617_{0.02}$ & 
  \cellcolor{green!8}$0.272_{0.14}$ & 
  \cellcolor{green!36}$0.462_{0.11}$ & 
  \cellcolor{green!62}\underline{$0.632_{0.02}$} \\
 & LIME& 
  \cellcolor{gray!100}$0.780_{0.24}$ & 
  \cellcolor{cyan!24}$0.338_{0.19}$ & 
  \cellcolor{cyan!92}$0.718_{0.02}$ & 
  \cellcolor{cyan!100}$0.760_{0.02}$ & 
  \cellcolor{orange!95}$0.814_{0.02}$ & 
  \cellcolor{orange!100}$0.837_{0.01}$ & 
  \cellcolor{orange!96}$0.823_{0.02}$ & 
  \cellcolor{green!26}$0.391_{0.27}$ & 
  \cellcolor{green!72}$0.702_{0.19}$ & 
  \cellcolor{green!100}\underline{$0.882_{0.02}$} \\
 & SHAP& 
  \cellcolor{gray!96}$0.765_{0.23}$ & 
  \cellcolor{cyan!23}$0.333_{0.19}$ & 
  \cellcolor{cyan!92}$0.718_{0.01}$ & 
  \cellcolor{cyan!98}$0.751_{0.01}$ & 
  \cellcolor{orange!95}$0.818_{0.02}$ & 
  \cellcolor{orange!99}$0.832_{0.03}$ & 
  \cellcolor{orange!96}$0.823_{0.03}$ & 
  \cellcolor{green!24}$0.378_{0.26}$ & 
  \cellcolor{green!69}$0.679_{0.19}$ & 
  \cellcolor{green!95}\underline{$0.853_{0.03}$} \\
 & Avg& 
  $0.713_{0.20}$ & 
  $0.301_{0.16}$ & 
  $0.654_{0.01}$ & 
  $0.698_{0.01}$ & 
  $0.750_{0.01}$ & 
  $0.765_{0.02}$ & 
  $0.758_{0.02}$ & 
  $0.344_{0.22}$ & 
  $0.610_{0.17}$ & 
  \underline{$0.786_{0.03}$} \\
\midrule
\multirow{5}{*}{SST2} & G& 
  \cellcolor{gray!28}\underline{$0.515_{0.18}$} & 
  \cellcolor{cyan!6}$0.240_{0.05}$ & 
  \cellcolor{cyan!27}$0.358_{0.14}$ & 
  \cellcolor{cyan!42}$0.441_{0.13}$ & 
  \cellcolor{orange!24}$0.489_{0.16}$ & 
  \cellcolor{orange!21}$0.477_{0.15}$ & 
  \cellcolor{orange!21}$0.475_{0.16}$ & 
  \cellcolor{green!3}$0.239_{0.06}$ & 
  \cellcolor{green!9}$0.277_{0.09}$ & 
  \cellcolor{green!26}$0.396_{0.18}$ \\
 & IG&   
  \cellcolor{gray!0}\underline{$0.410_{0.13}$} & 
  \cellcolor{cyan!0}$0.203_{0.04}$ & 
  \cellcolor{cyan!15}$0.287_{0.10}$ & 
  \cellcolor{cyan!27}$0.358_{0.10}$ & 
  \cellcolor{orange!3}$0.390_{0.12}$ & 
  \cellcolor{orange!0}$0.375_{0.11}$ & 
  \cellcolor{orange!1}$0.382_{0.11}$ & 
  \cellcolor{green!0}$0.217_{0.05}$ & 
  \cellcolor{green!4}$0.246_{0.07}$ & 
  \cellcolor{green!16}$0.325_{0.13}$ \\
 & LIME& 
  \cellcolor{gray!43}\underline{$0.573_{0.22}$} & 
  \cellcolor{cyan!2}$0.214_{0.08}$ & 
  \cellcolor{cyan!30}$0.375_{0.18}$ & 
  \cellcolor{cyan!50}$0.483_{0.17}$ & 
  \cellcolor{orange!35}$0.537_{0.20}$ & 
  \cellcolor{orange!31}$0.521_{0.19}$ & 
  \cellcolor{orange!31}$0.520_{0.19}$ & 
  \cellcolor{green!3}$0.237_{0.11}$ & 
  \cellcolor{green!10}$0.286_{0.15}$ & 
  \cellcolor{green!31}$0.430_{0.25}$ \\
 & SHAP& 
  \cellcolor{gray!42}\underline{$0.567_{0.22}$} & 
  \cellcolor{cyan!2}$0.215_{0.07}$ & 
  \cellcolor{cyan!30}$0.374_{0.19}$ & 
  \cellcolor{cyan!50}$0.484_{0.17}$ & 
  \cellcolor{orange!34}$0.535_{0.20}$ & 
  \cellcolor{orange!31}$0.523_{0.19}$ & 
  \cellcolor{orange!31}$0.518_{0.18}$ & 
  \cellcolor{green!2}$0.235_{0.11}$ & 
  \cellcolor{green!10}$0.285_{0.16}$ & 
  \cellcolor{green!31}$0.423_{0.25}$ \\
 & Avg& 
  \underline{$0.516_{0.19}$} & 
  $0.218_{0.06}$ & 
  $0.349_{0.15}$ & 
  $0.442_{0.14}$ & 
  $0.488_{0.17}$ & 
  $0.474_{0.16}$ & 
  $0.474_{0.16}$ & 
  $0.232_{0.08}$ & 
  $0.274_{0.12}$ & 
  $0.394_{0.20}$ \\
\midrule
\multirow{5}{*}{Trustpilot} & G& 
  \cellcolor{gray!25}$0.506_{0.17}$ & 
  \cellcolor{cyan!39}$0.421_{0.11}$ & 
  \cellcolor{cyan!51}$0.489_{0.17}$ & 
  \cellcolor{cyan!58}\underline{$0.531_{0.12}$} & 
  \cellcolor{orange!25}$0.491_{0.14}$ & 
  \cellcolor{orange!24}$0.487_{0.08}$ & 
  \cellcolor{orange!23}$0.483_{0.06}$ & 
  \cellcolor{green!15}$0.321_{0.14}$ & 
  \cellcolor{green!15}$0.318_{0.10}$ & 
  \cellcolor{green!43}$0.504_{0.16}$ \\
 & IG&   
  \cellcolor{gray!20}$0.488_{0.16}$ & 
  \cellcolor{cyan!38}$0.420_{0.10}$ & 
  \cellcolor{cyan!49}$0.477_{0.17}$ & 
  \cellcolor{cyan!55}\underline{$0.513_{0.12}$} & 
  \cellcolor{orange!17}$0.458_{0.13}$ & 
  \cellcolor{orange!17}$0.456_{0.07}$ & 
  \cellcolor{orange!16}$0.451_{0.05}$ & 
  \cellcolor{green!15}$0.320_{0.14}$ & 
  \cellcolor{green!14}$0.311_{0.10}$ & 
  \cellcolor{green!39}$0.480_{0.15}$ \\
 & LIME& 
  \cellcolor{gray!39}$0.558_{0.20}$ & 
  \cellcolor{cyan!44}$0.451_{0.12}$ & 
  \cellcolor{cyan!57}$0.521_{0.18}$ & 
  \cellcolor{cyan!65}\underline{$0.566_{0.13}$} & 
  \cellcolor{orange!35}$0.539_{0.17}$ & 
  \cellcolor{orange!36}$0.542_{0.10}$ & 
  \cellcolor{orange!35}$0.537_{0.06}$ & 
  \cellcolor{green!18}$0.343_{0.19}$ & 
  \cellcolor{green!20}$0.356_{0.15}$ & 
  \cellcolor{green!50}$0.555_{0.19}$ \\
 & SHAP& 
  \cellcolor{gray!38}$0.551_{0.19}$ & 
  \cellcolor{cyan!42}$0.443_{0.12}$ & 
  \cellcolor{cyan!56}$0.517_{0.18}$ & 
  \cellcolor{cyan!63}\underline{$0.558_{0.13}$} & 
  \cellcolor{orange!34}$0.534_{0.17}$ & 
  \cellcolor{orange!33}$0.530_{0.09}$ & 
  \cellcolor{orange!34}$0.534_{0.07}$ & 
  \cellcolor{green!16}$0.325_{0.19}$ & 
  \cellcolor{green!18}$0.340_{0.14}$ & 
  \cellcolor{green!50}$0.551_{0.19}$ \\
 & Avg& 
  $0.526_{0.18}$ & 
  $0.434_{0.11}$ & 
  $0.501_{0.18}$ & 
  \underline{$0.542_{0.13}$} & 
  $0.506_{0.15}$ & 
  $0.504_{0.09}$ & 
  $0.501_{0.06}$ & 
  $0.327_{0.16}$ & 
  $0.331_{0.12}$ & 
  $0.523_{0.17}$ \\
\bottomrule
\end{tabular}
}
\caption{Composite Scores (mean$_{\scriptstyle\text{std}}$) with $\alpha=0.25$ averaged over the five PLMs. \textit{Avg} refers to the mean and pooled standard deviation over the four explainers.}
\label{tab:scores_alpha_25_avg}
\end{table*}

\begin{table*}[ht]
\small
\scriptsize
\setlength{\tabcolsep}{3pt} %
\centering
\resizebox{0.7\linewidth}{!}{
\begin{tabular}{l l c c c c c c c c c c}
\toprule
Dataset & Expl & None & DPB500 & DPB1000 & DPB1500 & DPP118 & DPP137 & DPP165 & TEM1 & TEM2 & TEM3 \\
\midrule
\multirow{5}{*}{AG News} 
& G & \cellcolor{gray!83}$0.766_{0.13}$ & \cellcolor{cyan!5}$0.297_{0.19}$ & \cellcolor{cyan!83}$0.710_{0.01}$ & \cellcolor{cyan!92}$0.760_{0.01}$ & \cellcolor{orange!84}$0.811_{0.00}$ & \cellcolor{orange!87}$0.820_{0.01}$ & \cellcolor{orange!86}$0.818_{0.01}$ & \cellcolor{green!10}$0.343_{0.26}$ & \cellcolor{green!55}$0.623_{0.19}$ & \cellcolor{green!88}\underline{$0.828_{0.02}$} \\
 & IG & \cellcolor{gray!59}$0.696_{0.09}$ & \cellcolor{cyan!0}$0.270_{0.17}$ & \cellcolor{cyan!67}$0.624_{0.01}$ & \cellcolor{cyan!76}$0.674_{0.01}$ & \cellcolor{orange!56}$0.712_{0.01}$ & \cellcolor{orange!58}$0.719_{0.02}$ & \cellcolor{orange!57}$0.717_{0.01}$ & \cellcolor{green!3}$0.301_{0.21}$ & \cellcolor{green!41}$0.533_{0.15}$ & \cellcolor{green!73}\underline{$0.731_{0.02}$} \\
 & LIME & \cellcolor{gray!100}$0.816_{0.16}$ & \cellcolor{cyan!11}$0.332_{0.23}$ & \cellcolor{cyan!91}$0.753_{0.01}$ & \cellcolor{cyan!100}$0.797_{0.01}$ & \cellcolor{orange!95}$0.848_{0.01}$ & \cellcolor{orange!100}$0.864_{0.01}$ & \cellcolor{orange!97}$0.854_{0.01}$ & \cellcolor{green!16}$0.381_{0.29}$ & \cellcolor{green!66}$0.693_{0.20}$ & \cellcolor{green!100}\underline{$0.898_{0.02}$} \\
 & SHAP & \cellcolor{gray!96}$0.806_{0.15}$ & \cellcolor{cyan!11}$0.329_{0.22}$ & \cellcolor{cyan!91}$0.752_{0.01}$ & \cellcolor{cyan!98}$0.791_{0.01}$ & \cellcolor{orange!96}$0.851_{0.01}$ & \cellcolor{orange!99}$0.861_{0.02}$ & \cellcolor{orange!97}$0.854_{0.02}$ & \cellcolor{green!15}$0.372_{0.29}$ & \cellcolor{green!64}$0.678_{0.20}$ & \cellcolor{green!96}\underline{$0.878_{0.02}$} \\
 & Avg & $0.771_{0.13}$ & $0.307_{0.20}$ & $0.710_{0.01}$ & $0.756_{0.01}$ & $0.806_{0.01}$ & $0.816_{0.01}$ & $0.811_{0.01}$ & $0.349_{0.26}$ & $0.632_{0.19}$ & \underline{$0.834_{0.02}$} \\
\midrule
\multirow{5}{*}{SST2} & G & \cellcolor{gray!23}\underline{$0.592_{0.21}$} & \cellcolor{cyan!4}$0.294_{0.09}$ & \cellcolor{cyan!32}$0.439_{0.19}$ & \cellcolor{cyan!51}$0.541_{0.17}$ & \cellcolor{orange!21}$0.591_{0.20}$ & \cellcolor{orange!19}$0.583_{0.20}$ & \cellcolor{orange!18}$0.580_{0.20}$ & \cellcolor{green!2}$0.293_{0.10}$ & \cellcolor{green!10}$0.342_{0.13}$ & \cellcolor{green!32}$0.479_{0.23}$ \\
 & IG & \cellcolor{gray!0}$0.522_{0.19}$ & \cellcolor{cyan!0}$0.269_{0.08}$ & \cellcolor{cyan!23}$0.391_{0.16}$ & \cellcolor{cyan!41}$0.486_{0.15}$ & \cellcolor{orange!2}\underline{$0.525_{0.17}$} & \cellcolor{orange!0}$0.515_{0.16}$ & \cellcolor{orange!0}$0.518_{0.17}$ & \cellcolor{green!0}$0.278_{0.09}$ & \cellcolor{green!7}$0.322_{0.12}$ & \cellcolor{green!24}$0.432_{0.19}$ \\
 & LIME & \cellcolor{gray!36}\underline{$0.630_{0.24}$} & \cellcolor{cyan!1}$0.277_{0.10}$ & \cellcolor{cyan!34}$0.450_{0.21}$ & \cellcolor{cyan!56}$0.569_{0.20}$ & \cellcolor{orange!30}$0.623_{0.23}$ & \cellcolor{orange!27}$0.613_{0.22}$ & \cellcolor{orange!27}$0.610_{0.22}$ & \cellcolor{green!2}$0.292_{0.13}$ & \cellcolor{green!11}$0.349_{0.17}$ & \cellcolor{green!36}$0.501_{0.27}$ \\
 & SHAP & \cellcolor{gray!35}\underline{$0.627_{0.24}$} & \cellcolor{cyan!1}$0.277_{0.10}$ & \cellcolor{cyan!34}$0.449_{0.22}$ & \cellcolor{cyan!56}$0.570_{0.20}$ & \cellcolor{orange!30}$0.622_{0.23}$ & \cellcolor{orange!28}$0.614_{0.22}$ & \cellcolor{orange!26}$0.609_{0.22}$ & \cellcolor{green!1}$0.290_{0.13}$ & \cellcolor{green!11}$0.347_{0.18}$ & \cellcolor{green!35}$0.497_{0.28}$ \\
 & Avg & \underline{$0.593_{0.22}$} & $0.279_{0.09}$ & $0.432_{0.20}$ & $0.542_{0.18}$ & $0.590_{0.21}$ & $0.581_{0.20}$ & $0.579_{0.20}$ & $0.288_{0.11}$ & $0.340_{0.15}$ & $0.477_{0.24}$ \\
\midrule
\multirow{5}{*}{Trustpilot} & G & \cellcolor{gray!19}$0.579_{0.20}$ & \cellcolor{cyan!43}$0.501_{0.11}$ & \cellcolor{cyan!62}$0.600_{0.18}$ & \cellcolor{cyan!76}\underline{$0.671_{0.08}$} & \cellcolor{orange!25}$0.604_{0.17}$ & \cellcolor{orange!32}$0.628_{0.06}$ & \cellcolor{orange!34}$0.634_{0.04}$ & \cellcolor{green!20}$0.407_{0.14}$ & \cellcolor{green!22}$0.419_{0.13}$ & \cellcolor{green!56}$0.626_{0.19}$ \\
 & IG & \cellcolor{gray!15}$0.567_{0.19}$ & \cellcolor{cyan!43}$0.500_{0.11}$ & \cellcolor{cyan!61}$0.592_{0.18}$ & \cellcolor{cyan!73}\underline{$0.659_{0.08}$} & \cellcolor{orange!18}$0.581_{0.16}$ & \cellcolor{orange!26}$0.608_{0.05}$ & \cellcolor{orange!28}$0.613_{0.04}$ & \cellcolor{green!20}$0.406_{0.14}$ & \cellcolor{green!22}$0.415_{0.13}$ & \cellcolor{green!53}$0.610_{0.18}$ \\
 & LIME & \cellcolor{gray!31}$0.614_{0.21}$ & \cellcolor{cyan!47}$0.521_{0.12}$ & \cellcolor{cyan!66}$0.621_{0.19}$ & \cellcolor{cyan!80}\underline{$0.695_{0.09}$} & \cellcolor{orange!34}$0.636_{0.19}$ & \cellcolor{orange!42}$0.665_{0.07}$ & \cellcolor{orange!44}$0.670_{0.04}$ & \cellcolor{green!23}$0.422_{0.18}$ & \cellcolor{green!26}$0.445_{0.16}$ & \cellcolor{green!61}$0.660_{0.21}$ \\
 & SHAP & \cellcolor{gray!29}$0.610_{0.21}$ & \cellcolor{cyan!46}$0.515_{0.12}$ & \cellcolor{cyan!66}$0.619_{0.19}$ & \cellcolor{cyan!79}\underline{$0.689_{0.09}$} & \cellcolor{orange!33}$0.632_{0.19}$ & \cellcolor{orange!40}$0.657_{0.07}$ & \cellcolor{orange!43}$0.668_{0.05}$ & \cellcolor{green!21}$0.409_{0.17}$ & \cellcolor{green!25}$0.433_{0.15}$ & \cellcolor{green!61}$0.658_{0.21}$ \\
 & Avg & $0.593_{0.20}$ & $0.509_{0.12}$ & $0.608_{0.18}$ & \underline{$0.679_{0.09}$} & $0.613_{0.18}$ & $0.640_{0.06}$ & $0.646_{0.04}$ & $0.411_{0.16}$ & $0.428_{0.14}$ & $0.639_{0.19}$ \\
\bottomrule
\end{tabular}
}
\caption{Composite Scores (mean$_{\scriptstyle\text{std}}$) for $\alpha=0.5$ averaged over five PLMs. \textit{Avg} refers to the mean and pooled standard deviation over the four explainers.}
\label{tab:scores_alpha_50}
\end{table*}

\begin{table*}[ht!]
\small
\scriptsize
\setlength{\tabcolsep}{3pt} 
\centering
\resizebox{0.7\linewidth}{!}{
\begin{tabular}{l l c c c c c c c c c c}
\toprule
Dataset & Expl & None & DPB500 & DPB1000 & DPB1500 & DPP118 & DPP137 & DPP165 & TEM1 & TEM2 & TEM3 \\
\midrule
\multirow{4}{*}{AG News} & G & \cellcolor{gray!88}$0.827_{0.09}$ & \cellcolor{cyan!2}$0.308_{0.24}$ & \cellcolor{cyan!87}$0.766_{0.00}$ & \cellcolor{cyan!96}$0.815_{0.00}$ & \cellcolor{orange!88}$0.863_{0.00}$ & \cellcolor{orange!90}$0.869_{0.01}$ & \cellcolor{orange!90}$0.868_{0.01}$ & \cellcolor{green!3}$0.352_{0.30}$ & \cellcolor{green!54}$0.649_{0.21}$ & \cellcolor{green!93}\underline{$0.878_{0.01}$} \\
 & IG & \cellcolor{gray!72}$0.792_{0.08}$ & \cellcolor{cyan!0}$0.295_{0.23}$ & \cellcolor{cyan!79}$0.723_{0.01}$ & \cellcolor{cyan!88}$0.772_{0.01}$ & \cellcolor{orange!67}$0.814_{0.00}$ & \cellcolor{orange!69}$0.818_{0.01}$ & \cellcolor{orange!68}$0.817_{0.00}$ & \cellcolor{green!0}$0.331_{0.28}$ & \cellcolor{green!46}$0.604_{0.19}$ & \cellcolor{green!85}\underline{$0.830_{0.01}$} \\
 & LIME & \cellcolor{gray!100}$0.852_{0.10}$ & \cellcolor{cyan!5}$0.326_{0.26}$ & \cellcolor{cyan!91}$0.787_{0.01}$ & \cellcolor{cyan!100}$0.834_{0.01}$ & \cellcolor{orange!96}$0.882_{0.01}$ & \cellcolor{orange!100}$0.891_{0.01}$ & \cellcolor{orange!97}$0.886_{0.01}$ & \cellcolor{green!6}$0.371_{0.32}$ & \cellcolor{green!60}$0.684_{0.22}$ & \cellcolor{green!100}\underline{$0.913_{0.01}$} \\
 & SHAP & \cellcolor{gray!97}$0.847_{0.09}$ & \cellcolor{cyan!5}$0.324_{0.26}$ & \cellcolor{cyan!91}$0.787_{0.00}$ & \cellcolor{cyan!99}$0.830_{0.01}$ & \cellcolor{orange!96}$0.883_{0.01}$ & \cellcolor{orange!99}$0.889_{0.01}$ & \cellcolor{orange!97}$0.886_{0.01}$ & \cellcolor{green!6}$0.367_{0.32}$ & \cellcolor{green!59}$0.676_{0.21}$ & \cellcolor{green!98}\underline{$0.904_{0.01}$} \\
 & Avg & $0.830_{0.09}$ & $0.313_{0.25}$ & $0.766_{0.01}$ & $0.813_{0.01}$ & $0.861_{0.01}$ & $0.867_{0.01}$ & $0.864_{0.01}$ & $0.355_{0.30}$ & $0.653_{0.21}$ & \underline{$0.881_{0.01}$} \\
\midrule
\multirow{4}{*}{SST2} & G & \cellcolor{gray!16}$0.669_{0.25}$ & \cellcolor{cyan!9}$0.347_{0.12}$ & \cellcolor{cyan!41}$0.520_{0.23}$ & \cellcolor{cyan!64}$0.641_{0.21}$ & \cellcolor{orange!16}\underline{$0.693_{0.24}$} & \cellcolor{orange!15}$0.689_{0.24}$ & \cellcolor{orange!13}$0.685_{0.24}$ & \cellcolor{green!2}$0.347_{0.14}$ & \cellcolor{green!13}$0.408_{0.17}$ & \cellcolor{green!39}$0.562_{0.28}$ \\
 & IG & \cellcolor{gray!0}$0.634_{0.24}$ & \cellcolor{cyan!7}$0.335_{0.12}$ & \cellcolor{cyan!37}$0.496_{0.22}$ & \cellcolor{cyan!59}$0.613_{0.20}$ & \cellcolor{orange!2}\underline{$0.660_{0.23}$} & \cellcolor{orange!0}$0.656_{0.22}$ & \cellcolor{orange!0}$0.654_{0.22}$ & \cellcolor{green!1}$0.339_{0.13}$ & \cellcolor{green!11}$0.397_{0.16}$ & \cellcolor{green!35}$0.538_{0.26}$ \\
 & LIME & \cellcolor{gray!24}$0.688_{0.26}$ & \cellcolor{cyan!8}$0.339_{0.13}$ & \cellcolor{cyan!42}$0.525_{0.25}$ & \cellcolor{cyan!66}$0.655_{0.22}$ & \cellcolor{orange!23}\underline{$0.709_{0.25}$} & \cellcolor{orange!21}$0.704_{0.25}$ & \cellcolor{orange!19}$0.700_{0.25}$ & \cellcolor{green!2}$0.346_{0.15}$ & \cellcolor{green!13}$0.411_{0.19}$ & \cellcolor{green!41}$0.573_{0.30}$ \\
 & SHAP & \cellcolor{gray!23}$0.686_{0.26}$ & \cellcolor{cyan!8}$0.339_{0.13}$ & \cellcolor{cyan!42}$0.525_{0.25}$ & \cellcolor{cyan!66}$0.655_{0.22}$ & \cellcolor{orange!23}\underline{$0.708_{0.25}$} & \cellcolor{orange!21}$0.705_{0.25}$ & \cellcolor{orange!19}$0.699_{0.25}$ & \cellcolor{green!2}$0.345_{0.15}$ & \cellcolor{green!13}$0.410_{0.19}$ & \cellcolor{green!41}$0.571_{0.30}$ \\
 & Avg & $0.669_{0.25}$ & $0.340_{0.12}$ & $0.517_{0.24}$ & $0.641_{0.21}$ & \underline{$0.693_{0.24}$} & $0.689_{0.24}$ & $0.685_{0.24}$ & $0.344_{0.14}$ & $0.407_{0.18}$ & $0.561_{0.28}$ \\
\midrule
\multirow{4}{*}{Trustpilot} & G & \cellcolor{gray!8}$0.653_{0.24}$ & \cellcolor{cyan!53}$0.581_{0.12}$ & \cellcolor{cyan!77}$0.711_{0.20}$ & \cellcolor{cyan!95}\underline{$0.812_{0.04}$} & \cellcolor{orange!26}$0.716_{0.20}$ & \cellcolor{orange!48}$0.770_{0.04}$ & \cellcolor{orange!55}$0.786_{0.03}$ & \cellcolor{green!27}$0.492_{0.15}$ & \cellcolor{green!32}$0.520_{0.16}$ & \cellcolor{green!71}$0.748_{0.22}$ \\
 & IG & \cellcolor{gray!6}$0.647_{0.23}$ & \cellcolor{cyan!52}$0.580_{0.12}$ & \cellcolor{cyan!76}$0.707_{0.19}$ & \cellcolor{cyan!94}\underline{$0.806_{0.04}$} & \cellcolor{orange!21}$0.705_{0.19}$ & \cellcolor{orange!44}$0.759_{0.04}$ & \cellcolor{orange!51}$0.775_{0.03}$ & \cellcolor{green!27}$0.492_{0.15}$ & \cellcolor{green!32}$0.518_{0.16}$ & \cellcolor{green!70}$0.740_{0.21}$ \\
 & LIME & \cellcolor{gray!16}$0.670_{0.24}$ & \cellcolor{cyan!54}$0.591_{0.13}$ & \cellcolor{cyan!79}$0.722_{0.20}$ & \cellcolor{cyan!98}\underline{$0.824_{0.05}$} & \cellcolor{orange!33}$0.732_{0.21}$ & \cellcolor{orange!56}$0.788_{0.05}$ & \cellcolor{orange!63}$0.804_{0.03}$ & \cellcolor{green!28}$0.500_{0.17}$ & \cellcolor{green!34}$0.533_{0.18}$ & \cellcolor{green!74}$0.765_{0.23}$ \\
 & SHAP & \cellcolor{gray!15}$0.668_{0.24}$ & \cellcolor{cyan!54}$0.588_{0.13}$ & \cellcolor{cyan!79}$0.721_{0.20}$ & \cellcolor{cyan!97}\underline{$0.821_{0.05}$} & \cellcolor{orange!32}$0.731_{0.21}$ & \cellcolor{orange!54}$0.784_{0.04}$ & \cellcolor{orange!62}$0.803_{0.03}$ & \cellcolor{green!27}$0.494_{0.16}$ & \cellcolor{green!33}$0.527_{0.17}$ & \cellcolor{green!74}$0.764_{0.22}$ \\
 & Avg & $0.660_{0.24}$ & $0.585_{0.12}$ & $0.715_{0.20}$ & \underline{$0.816_{0.04}$} & $0.721_{0.20}$ & $0.775_{0.04}$ & $0.792_{0.03}$ & $0.495_{0.16}$ & $0.525_{0.17}$ & $0.754_{0.22}$ \\
\bottomrule
\end{tabular}
}
\caption{Composite Scores (mean$_{\scriptstyle\text{std}}$) for $\alpha=0.75$ averaged over five PLMs. \textit{Avg} refers to the mean and pooled standard deviation over the four explainers.}
\label{tab:scores_alpha_75}
\end{table*} 

\begin{figure}[ht!]
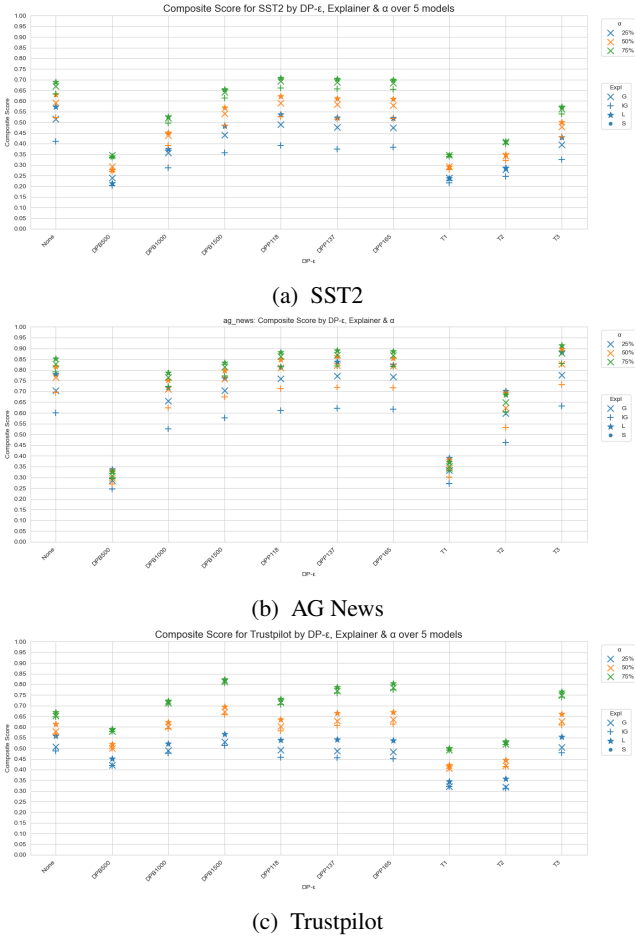

    \centering
    \begin{subfigure}{\linewidth}
        \includegraphics[width=\linewidth]{figures/scores_each_explainer_all_alpha_sst2.pdf}
        \caption{\label{subfig_sst2} SST2}
        \label{fig:alpha_25_avg_sst2}
    \end{subfigure}
    \hfill
    \begin{subfigure}{\linewidth}
        \includegraphics[width=\linewidth]{figures/scores_each_explainer_all_alpha_ag_news.pdf}
        \caption{\label{subfig_ag_news} AG News}
        \label{fig:alpha_25_avg_ag_news}
    \end{subfigure}
    \hfill
    \begin{subfigure}{\linewidth}
        \includegraphics[width=\linewidth]{figures/scores_each_explainer_all_alpha_trustpilot.pdf}
        \caption{\label{subfig_trustpilot} Trustpilot}
        \label{fig:alpha_25_avg_trsutpilot}
    \end{subfigure}
    \caption{Composite Score by DP-$\varepsilon$, explainer, $\alpha$, dataset over 5 models.}
    \label{fig:all_fig_all_datasets_all_alpha}
\end{figure}

To better visualize the results from Tables~\ref{tab:scores_alpha_25_avg}, \ref{tab:scores_alpha_50}, and \ref{tab:scores_alpha_75} for each of the three datasets, we present corresponding plots in Figures~\ref{fig:alpha_25_avg_sst2}, \ref{fig:alpha_25_avg_ag_news}, and \ref{fig:alpha_25_avg_trsutpilot}. Each figure corresponds to one dataset, and each data point represents the composite score of a (DP-$\varepsilon$, explainer, $\alpha$) triplet, averaged over the five PLMs. In these plots, different shapes indicate different explainers, while colors denote the $\alpha$ values \textit{(0.25, 0.5, 0.75)} where explainers are acronymed as follows: (G:Gradient, IG:Integrated Gradient, L:LIME, S:SHAP). 

In Table \ref{tab:consolidated_avg_scores_color}, we present composite score values, but we consolidate the composite score values averaged over the four explainers (instead of per explainer as done previously) for each (dataset, $\alpha$, DP-$\varepsilon$) triplet thus allowing us to compare and evaluate how each DP-$\varepsilon$ combination affect the composite scores over all the explainers allowing us to detect the \textit{``sweet spots''}
for each $\alpha$ value/scenario in each dataset.

\begin{table*}[ht]
\small
\scriptsize
\setlength{\tabcolsep}{3pt} 
\centering
\resizebox{0.7\linewidth}{!}{
\begin{tabular}{l l l c c c c c c c c c c}
\toprule
Dataset & Expl & \textbf{$\alpha$} & None & DPB500 & DPB1000 & DPB1500 & DPP118 & DPP137 & DPP165 & TEM1 & TEM2 & TEM3 \\
\midrule
\multirow{3}{*}{AG News} & Avg & 0.25 & \cellcolor{gray!62}$0.713_{0.20}$ & \cellcolor{cyan!17}$0.301_{0.16}$ & \cellcolor{cyan!88}$0.654_{0.01}$ & \cellcolor{cyan!95}$0.698_{0.01}$ & \cellcolor{orange!70}$0.750_{0.01}$ & \cellcolor{orange!74}$0.765_{0.02}$ & \cellcolor{orange!72}$0.758_{0.02}$ & \cellcolor{green!17}$0.344_{0.22}$ & \cellcolor{green!60}$0.610_{0.17}$ & \cellcolor{green!85}\underline{$0.786_{0.03}$} \\
 & Avg & 0.50 & \cellcolor{gray!82}$0.771_{0.13}$ & \cellcolor{cyan!18}$0.307_{0.20}$ & \cellcolor{cyan!96}$0.710_{0.01}$ & \cellcolor{cyan!100}$0.756_{0.01}$ & \cellcolor{orange!84}$0.806_{0.01}$ & \cellcolor{orange!87}$0.816_{0.01}$ & \cellcolor{orange!85}$0.811_{0.01}$ & \cellcolor{green!18}$0.349_{0.26}$ & \cellcolor{green!62}$0.632_{0.19}$ & \cellcolor{green!91}\underline{$0.834_{0.02}$} \\
 & Avg & 0.75 & \cellcolor{gray!100}$0.830_{0.09}$ & \cellcolor{cyan!19}$0.313_{0.25}$ & \cellcolor{cyan!100}$0.766_{0.01}$ & \cellcolor{cyan!100}$0.813_{0.01}$ & \cellcolor{orange!98}$0.861_{0.01}$ & \cellcolor{orange!100}$0.867_{0.01}$ & \cellcolor{orange!99}$0.864_{0.01}$ & \cellcolor{green!19}$0.355_{0.30}$ & \cellcolor{green!65}$0.653_{0.21}$ & \cellcolor{green!100}\underline{$0.881_{0.01}$} \\
\midrule
\multirow{3}{*}{SST2} & Avg & 0.25 & \cellcolor{gray!0}\underline{$0.516_{0.19}$} & \cellcolor{cyan!0}$0.218_{0.06}$ & \cellcolor{cyan!26}$0.349_{0.15}$ & \cellcolor{cyan!45}$0.442_{0.14}$ & \cellcolor{orange!47}$0.488_{0.17}$ & \cellcolor{orange!43}$0.474_{0.16}$ & \cellcolor{orange!43}$0.474_{0.16}$ & \cellcolor{green!0}$0.232_{0.08}$ & \cellcolor{green!6}$0.274_{0.12}$ & \cellcolor{green!24}$0.394_{0.20}$ \\
 & Avg & 0.50 & \cellcolor{gray!25}\underline{$0.593_{0.22}$} & \cellcolor{cyan!12}$0.279_{0.09}$ & \cellcolor{cyan!43}$0.432_{0.20}$ & \cellcolor{cyan!65}$0.542_{0.18}$ & \cellcolor{orange!61}$0.590_{0.21}$ & \cellcolor{orange!58}$0.581_{0.20}$ & \cellcolor{orange!57}$0.579_{0.20}$ & \cellcolor{green!8}$0.288_{0.11}$ & \cellcolor{green!16}$0.340_{0.15}$ & \cellcolor{green!37}$0.477_{0.24}$ \\
 & Avg & 0.75 & \cellcolor{gray!48}$0.669_{0.25}$ & \cellcolor{cyan!24}$0.340_{0.12}$ & \cellcolor{cyan!62}$0.517_{0.24}$ & \cellcolor{cyan!83}$0.641_{0.21}$ & \cellcolor{orange!77}\underline{$0.693_{0.24}$} & \cellcolor{orange!76}$0.689_{0.24}$ & \cellcolor{orange!75}$0.685_{0.24}$ & \cellcolor{green!17}$0.344_{0.14}$ & \cellcolor{green!26}$0.407_{0.18}$ & \cellcolor{green!51}$0.561_{0.28}$ \\
\midrule
\multirow{3}{*}{Trustpilot} & Avg & 0.25 & \cellcolor{gray!32}$0.526_{0.18}$ & \cellcolor{cyan!44}$0.434_{0.11}$ & \cellcolor{cyan!58}$0.501_{0.18}$ & \cellcolor{cyan!65}\underline{$0.542_{0.13}$} & \cellcolor{orange!50}$0.506_{0.15}$ & \cellcolor{orange!49}$0.504_{0.09}$ & \cellcolor{orange!49}$0.501_{0.06}$ & \cellcolor{green!14}$0.327_{0.16}$ & \cellcolor{green!15}$0.331_{0.12}$ & \cellcolor{green!45}$0.523_{0.17}$ \\
 & Avg & 0.50 & \cellcolor{gray!51}$0.593_{0.20}$ & \cellcolor{cyan!60}$0.509_{0.12}$ & \cellcolor{cyan!74}$0.608_{0.18}$ & \cellcolor{cyan!88}\underline{$0.679_{0.09}$} & \cellcolor{orange!63}$0.613_{0.18}$ & \cellcolor{orange!68}$0.640_{0.06}$ & \cellcolor{orange!70}$0.646_{0.04}$ & \cellcolor{green!27}$0.411_{0.16}$ & \cellcolor{green!30}$0.428_{0.14}$ & \cellcolor{green!63}$0.639_{0.19}$ \\
 & Avg & 0.75 & \cellcolor{gray!70}$0.660_{0.24}$ & \cellcolor{cyan!73}$0.585_{0.12}$ & \cellcolor{cyan!98}$0.715_{0.20}$ & \cellcolor{cyan!100}\underline{$0.816_{0.04}$} & \cellcolor{orange!82}$0.721_{0.20}$ & \cellcolor{orange!89}$0.775_{0.04}$ & \cellcolor{orange!94}$0.792_{0.03}$ & \cellcolor{green!40}$0.495_{0.16}$ & \cellcolor{green!45}$0.525_{0.17}$ & \cellcolor{green!79}$0.754_{0.22}$ \\
\bottomrule
\end{tabular}
}
\caption{Consolidated Composite Scores (mean$_{\scriptstyle\text{std}}$) for three $\alpha$ values, showing \textbf{the average over four explainers}.}
\label{tab:consolidated_avg_scores_color}
\end{table*}



In Table~\ref{tab:base_large_scores}, we investigate the effect of model size on the trade-off between privacy, utility, and explainability; in particular, whether this trade-off differs between smaller and larger models. To this end, we average results over the four explainers, but instead of aggregating across all five models, we group the results by model size. The \textit{base} group includes \textsc{BERT-base} and \textsc{RoBERTa-base}, while the \textit{large} group includes their corresponding larger variants.

\begin{table}[ht!]
\centering
\small
\scriptsize
\setlength{\tabcolsep}{4pt}
\begin{tabular}{c l c c c}
\toprule
$\alpha$ & Dataset & Base & Large & $\Delta$ (Large--Base) \\
\midrule
\multirow{3}{*}{0.25} & AG News & $0.690_{0.15}$ & $0.566_{0.25}$ & $-0.125$ \\
                      & SST2 & $0.461_{0.13}$ & $0.273_{0.10}$ & $-0.188$ \\
                      & Trustpilot & $0.499_{0.11}$ & $0.380_{0.09}$ & $-0.119$ \\
\midrule
\multirow{3}{*}{0.50} & AG News & $0.736_{0.14}$ & $0.604_{0.29}$ & $-0.132$ \\
                      & SST2 & $0.568_{0.13}$ & $0.331_{0.12}$ & $-0.237$ \\
                      & Trustpilot & $0.621_{0.11}$ & $0.479_{0.12}$ & $-0.142$ \\
\midrule
\multirow{3}{*}{0.75} & AG News & $0.782_{0.13}$ & $0.643_{0.33}$ & $-0.140$ \\
                      & SST2 & $0.676_{0.14}$ & $0.390_{0.14}$ & $-0.286$ \\
                      & Trustpilot & $0.743_{0.12}$ & $0.578_{0.16}$ & $-0.165$ \\
\bottomrule
\end{tabular}
\caption{Comparison of average composite scores (mean$_{\scriptstyle std}$) between base and large models, averaged across DP-$\epsilon$ values for each dataset and $\alpha$.}
\label{tab:base_large_scores} 
\end{table}

\subsection{Composite Scores for Each Explainer Across DP-Methods and Datasets}
Tables \ref{tab:scores_alpha_25_avg}, \ref{tab:scores_alpha_50}, and \ref{tab:scores_alpha_75} facilitates checking, per each $\alpha$ values and dataset, and for each explainer, the DP-$\varepsilon$ method that provide the highest composite score, as well as for each DP-method, the $\varepsilon$ values that leads to the highest score for a specific DP method. In addition, we look at the plots in Figure 1 and based on our results from the composite score plots presented there for SST2, AG News, and Trustpilot, evaluated across different DP mechanisms (\(\varepsilon\)), four explainability methods, and utility–explanation trade-offs ($\alpha$ = 25\%, 50\%, 75\%), we can report the following:

\paragraph{General Observations.}  
Higher $\alpha$ values, which place greater emphasis on model utility, consistently result in higher composite scores across all datasets and DP configurations. Among the explainers, LIME and SHAP generally outperform Gradient and IG in the composite scores, particularly under strong privacy constraints. As expected, composite scores decline significantly in stricter DP settings (e.g., DPB500, TEM1), reflecting the trade-off between privacy and both utility and explanation quality.
 
\textbf{SST2} appears to be highly sensitive to DP noise, exhibiting considerable performance degradation under low privacy budgets. G and IG are especially affected in these settings, whereas LIME and SHAP demonstrate more stable performance, even under moderate DP configurations such as DPB1000 and \textsc{DP-Prompt}. Among the \textsc{TEM} methods, TEM3 shows partial recovery in performance, especially when greater weight is given to utility ($\alpha$ high).

On the other hand, \textbf{AG News}  displays stronger resilience to the effects of DP noise compared to SST2. LIME and SHAP maintain superior performance across most DP levels, with IG improving noticeably as \(\varepsilon\) increases. In particular, IG becomes competitive in configurations such as \textsc{DP-Prompt} and TEM3. The TEM3 variant performs especially well at ($\alpha$ = 75\%), indicating a favorable trade-off between utility and explanation quality.

\textbf{Trustpilot} demonstrates intermediate sensitivity to DP noise, falling between SST2 and AG News. Composite scores improve progressively with increasing \(\varepsilon\), and LIME remains the most effective explainer across the majority of settings. \textsc{DP-BART} and \textsc{DP-Prompt} mechanisms, particularly with moderate to high privacy budgets, offer reliable performance in maintaining explanation quality.

\paragraph{Cross-Dataset Trends.} Across all datasets, LIME and SHAP consistently outperform IG and Gradient, which are more affected by strong DP. Gradient is the most negatively impacted explainer, particularly in low-\(\varepsilon\) regimes and on the SST2 dataset. Among the DP mechanisms, \textit{DP-BART}-1500 and \textsc{DP-Prompt}-165 offer favorable trade-offs between privacy, utility, and explainability. While TEM1 and TEM2 generally result in considerable degradation, TEM3 performs better, especially for AG News and Trustpilot.

\begin{table}[ht!]
\centering
\small
\resizebox{0.99\linewidth}{!}{

\begin{tabular}{l 
                cc  
                cc 
                cc} 
\toprule
\multirow{2}{*}{\textbf{Privacy Tier}}
  & \multicolumn{2}{c}{\textbf{$\alpha=0.25$}}
  & \multicolumn{2}{c}{\textbf{$\alpha=0.50$}}
  & \multicolumn{2}{c}{\textbf{$\alpha=0.75$}} \\
\cmidrule(lr){2-3} \cmidrule(lr){4-5} \cmidrule(lr){6-7}
  & Expl–Data & Data(avg)
  & Expl–Data & Data(avg)
  & Expl–Data & Data(avg) \\
\midrule
None
  & \shortstack{LIME–AG News\\0.780}
  & \shortstack{AG News\\0.713}
  & \shortstack{LIME–AG News\\0.816}
  & \shortstack{AG News\\0.771}
  & \shortstack{LIME–AG News\\0.852}
  & \shortstack{AG News\\0.830}
  \\

Small $\varepsilon$
  & \shortstack{SHAP–AG News\\0.818}
  & \shortstack{Trustpilot\\0.434}
  & \shortstack{SHAP–AG News\\0.851}
  & \shortstack{Trustpilot\\0.509}
  & \shortstack{SHAP–AG News\\0.883}
  & \shortstack{Trustpilot\\0.585}
  \\

Medium $\varepsilon$
  & \shortstack{LIME–AG News\\0.837}
  & \shortstack{AG News\\0.654}
  & \shortstack{LIME–AG News\\0.864}
  & \shortstack{AG News\\0.710}
  & \shortstack{LIME–AG News\\0.891}
  & \shortstack{AG News\\0.766}
  \\

Large $\varepsilon$
  & \shortstack{LIME–AG News\\0.882}
  & \shortstack{AG News\\0.786}
  & \shortstack{LIME–AG News\\0.898}
  & \shortstack{AG News\\0.756}
  & \shortstack{LIME–AG News\\0.913}
  & \shortstack{AG News\\0.813}
  \\
\bottomrule
\end{tabular}

}

\caption{``Sweet spots'' summary across privacy tiers and $\alpha$ values. Each row corresponds to a privacy tier (None = no DP; Small/Medium/Large $\varepsilon$ = decreasing privacy). For each $\alpha$, the left column shows the best Explainer–Dataset pair (from Tables 1–3), and the right column shows the best Dataset based on explainer-averaged scores (from Table 5). Epsilon levels group as: Small (DPB500, DPP118, TEM1), Medium (DPB1000, DPP137, TEM2), and Large (DPB1500, DPP165, TEM3).}
\label{tab:sweet_spots_over_all_alphas_epsilon}
\end{table}

\subsection{ Comparison Over All Explainers Across DP-Methods and Datasets}

Table \ref{tab:consolidated_avg_scores_color} displays the average composite scores across the four explainers for each combination of dataset, $\alpha$ value, and DP method (DP-$\varepsilon$), averaged over the five employed PLMs. We draw the following results and insights.

\paragraph{Effect of the Utility–Explanation Trade-Off (\boldmath$\alpha$).}  
Across all datasets and DP setups, higher values of $\alpha$ (i.e., greater weight placed on utility) consistently yield higher composite scores. This is especially visible in the blue curves ($\alpha$ = 0.75), which dominate across most configurations. This trend reflects the stabilizing role of utility in the composite score, particularly under privacy-induced degradation.

\paragraph{Comparing DP Mechanisms.}  
Among the DP strategies, \textsc{DP-BART} and \textsc{DP-Prompt} (particularly at higher $\varepsilon$ values such as 1500 and 165) achieve the highest average composite scores, especially for AG News and Trustpilot. In contrast, the \textsc{TEM} methods (T1 and T2) result in a substantial performance drop, most notably for \textit{SST2}. As expected, DPB500 (strongest privacy budget) leads to the lowest performance across all datasets and $\alpha$ values, confirming the cost of tight privacy constraints.

\paragraph{Dataset Sensitivity.}  
SST2 consistently produces the lowest composite scores, reflecting its greater vulnerability to privacy-preserving perturbations. This could be due to shorter input lengths or more subtle semantic cues required for sentiment analysis. 
In contrast, AG News emerges as the most robust dataset, maintaining high composite scores even under moderate DP conditions. \textit{Trustpilot} performs moderately well and demonstrates steady recovery with increasing $\varepsilon$, particularly under \textsc{DP-BART} and \textsc{DP-Prompt}. 

\paragraph{Privacy–Performance Trade-Off.}  
Overall, there is a clear upward trend in composite scores with increasing $\varepsilon$. This pattern is most pronounced in the \textsc{DP-BART} and \textsc{DP-Prompt} curves, where performance improves steadily from $\varepsilon = 500$ to $\varepsilon = 1500/165$. This monotonic behavior confirms the expected trade-off: as privacy constraints are relaxed, utility and explanation quality are jointly improved.


\subsection{Identifying Sweet Spots for Privacy-Utility-Explainability}


We construct Table 7 using the results form  Tables 1,2,3 and 4, where we identify and present the sweet spots based on the composite scores across privacy tiers and alpha values settings for two cases: explainer-dataset pairs (from Table 1,2,3) and the datasets with the highest average scores over the four explainers (From Table 3). Based on these result in Table 7, we observe the following: 

\paragraph{Explainer–Dataset Consistency.}  
Across all \textit{privacy tiers}, LIME–AG News emerges as the top explainer–dataset pair in every case except the strictest privacy setting (\(\varepsilon\) small), where SHAP–AG News slightly outperforms it. This suggests that SHAP’s token‐level importance is more robust under heavy DP noise, but LIME generally provides the most faithful and useful explanations once privacy is relaxed.

\paragraph{Dataset‐Level Robustness.}  
When averaging across explainers, AG News consistently wins for no DP, medium privacy, and low privacy budgets. Under the tightest DP budget (small \(\varepsilon\)), however, Trustpilot takes the lead for all $\alpha$ values, indicating that explanations on Trustpilot degrade less, on average, under stringent privacy constraints.

\paragraph{Effect of Increasing \(\varepsilon\).}  
As \(\varepsilon\) increases (privacy is relaxed), composite scores rise monotonically for both the best explainer–dataset pair and the dataset‐average. The largest recovery occurs between small and medium \(\varepsilon\), especially for SHAP on AG News, highlighting that a moderate privacy budget recovers most of the lost explanation signal.

\paragraph{Utility-Explainability Trade‐Off ($\alpha$) Trends.}  
Increasing $\alpha$ from 0.25 (explanation‐focused) to 0.50 (balanced) to 0.75 (utility‐focused) uniformly boosts all composite scores, since more weight is placed on classification F1. Notably, the ranking of sweet spots remains stable: LIME–AG News dominates once \(\varepsilon\geq\) medium values, and \textit{Trustpilot} remains the dataset of choice under the smallest $\varepsilon$.

Although high composite scores are expected with large $\varepsilon$ values, it is noteworthy that even with medium and small $\varepsilon$ (i.e., stricter privacy), some explainers still achieve strong performance. These results are promising, demonstrating that explainability can be achieved under tighter privacy constraints. For example, SHAP achieves this across all three $\alpha$ values representing different utility–explanation trade-offs. This indicates that for certain combinations of dataset, explainer, and DP method, it is possible to attain high privacy while still maintaining strong utility and explanation quality, depending on the specific objective (i.e., lower or higher $\alpha$).

\subsection{Comparison Across Model Size}

In Table~\ref{tab:base_large_scores}, we investigate the effect of model size on the trade-off between privacy, utility, and explainability; in particular, whether this trade-off differs between smaller and larger models. To this end, we average results over the four explainers, but instead of aggregating across all five models (as in the previous tables), we group the results by model size. The \textit{base} group includes \textsc{BERT-base} and \textsc{RoBERTa-base}, while the \textit{large} group includes their corresponding larger variants. We exclude DeBERTa from this analysis, as we don't employ DeBERTa-large in our experiments. We consider this comparison between base and large variants of BERT and RoBERTa sufficient for a preliminary assessment of model size effects. Table ~\ref{tab:base_large_scores} presents a comparison of (average across explainers) composite scores (mean$_{\scriptstyle std}$) between base and large models, averaged across DP-$\epsilon$ values for each dataset and $\alpha$. We learn the following:

\paragraph{Base Models Outperform Large Models.}  
Across all datasets and $\alpha$ values, base models consistently outperform large models in terms of composite score. This is evident from the uniformly negative \(\Delta\) (Large – Base) values, ranging from \(-0.119\) to \(-0.286\). The results suggest that increasing model size under DP leads to a decline in the quality of the resulting predictions and explanations.

\paragraph{Dataset Sensitivity.}  
The \textit{SST2} dataset shows the largest negative impact of model size. For instance, the gap reaches \(\Delta = -0.286\) at \(\alpha = 0.75\). In contrast, the performance drop for \textit{AG News} and \textit{Trustpilot} remains more moderate (around \(-0.12\) to \(-0.16\)). This indicates that \textit{SST2} may be more sensitive to privacy-induced noise, possibly due to shorter inputs or the nature of sentiment-based classification.

\paragraph{Effect of $\alpha$ (Utility–Explanation Trade-Off).}  
The negative performance gap between large and base models generally widens as $\alpha$ increases (i.e., as more emphasis is placed on utility over explanation quality). This trend is particularly strong for \textit{SST2}, where the difference grows from \(-0.188\) (\(\alpha=0.25\)) to \(-0.286\) (\(\alpha=0.75\)). This suggests that utility suffers more under large models when trained with DP.

\paragraph{Stability and Variability.}  
The standard deviations are generally higher for large models, especially under \(\alpha = 0.75\). This points to greater instability and inconsistency in the performance of large models under DP, in addition to their lower mean scores.



\section{Discussion}
We reflect on the main findings of our experiments, giving way to a collection of practical recommendations, as well as future points to consider.


\paragraph{Does dataset matter in the privacy-explainability trade-off?}
An initial review of the experimental results suggests that the \textit{nature} of the dataset may contribute to different outcomes in terms of privacy versus explainability. One possible explanation for AG News outperforming SST2 and Trustpilot in terms of composite scores under multiple DP methods is that its topic‐classification nature results in inputs that contain multiple contextually aligned keywords, which are often robust to DP‐induced perturbations whereas SST2's short sentiment sentences and Trustpilot's informal user reviews may lose critical cues more easily. Additionally, PLMs achieve a higher baseline F1 on AG News, which could further buffer performance degradation under privacy constraints. Finally, the concentration of attribution weights on a few salient words in AG News may yield more faithful explanations compared to the more diffuse attributions required for sentiment or user-generated text. While these factors require further investigation, they offer a plausible account of AG News resilience in composite utility–explainability metrics with strict privacy budgets.

Despite the absolute differences exhibited by the AG News results in comparison to our other two datasets, one promising trend emerges. When looking at the composite score curves for any dataset in Figure \ref{fig:all_fig_all_datasets_all_alpha}
one can observe that a similar line is followed regardless of dataset, for any DP method or explainer. While these curves are not strictly uniform across datasets, the similar trends indicate a crucial point for future investigations, namely to determine the extent to which dataset matters in the combined study of privacy, utility, and explainability. Nevertheless, one must also keep in mind the differing performances of the various explainers we employ, which also seem to be impacted by the nature of dataset, and accordingly, the downstream task.

\paragraph{When privacy meets explainability.}
Harmonizing all of the experimental results, we converge on one important discussion point, which relates back to the initial research question we posed in this work: \textit{how does DP impact explainability?} Diving deeper, we also reflect on the question of whether DP and explainability can co-exist, or if there possibly remains friction between these two important mandates.

A helpful starting point lies in the comparison of baseline results (i.e., those achieved without any DP) and those post-privatization. While in the majority of cases, the application of DP leads to decreases in explainability, this is not always the case. Indeed, in some scenarios for all three employed DP methods, some results outperform the non-private baselines, and this occurs at least once among all (DP method, $\varepsilon$) combinations for each explainer. Furthermore, this result is most pronounced in the setting where explainability is most preferred ($\alpha = 0.75$), suggesting that when utility loss is less important, the effects of DP gain a more positive light. In this, we uncover that in some scenarios, DP actually serves to \textit{improve} explainability, a very promising prospect.

Beyond these results, we also wish to answer our research question from the angle of \textit{which} DP methods have \textit{which} impacts on downstream explainability. This must be approached carefully, as noted previously, since we cannot draw conclusions between DP mechanisms operating on different lexical levels and with different privacy guarantees. However, important trends do appear \textit{within} mechanisms, such as the relative stability of the \textsc{DP-Prompt} results as opposed to the much steeper fluctuations from \textsc{TEM} or \textsc{DP-BART}. We also conjecture that the choice of DP method also is tightly intertwined with the nature of the classification task, where generative methods (\textsc{DP-Prompt} and \textsc{DP-BART}) generally lead to more favorable results in our two binary classification tasks, whereas \textsc{TEM} consistently outperforms in the four-class AG News task. These results point to the importance of DP method choice, which can be heavily reliant on the downstream task at hand. This is made especially evident in Table \ref{tab:consolidated_avg_scores_color}, where all DP methods achieve the best consolidated score in at least one configuration.

\paragraph{Recommendations for Practitioners.}
Based on our findings, we compile a collection of recommendations for practitioners at the intersection of data privacy and explainability.

Based on our ``sweet spot'' analysis, for \textit{explanation quality} (low $\alpha$), practitioners can use SHAP at the strictest privacy setting, otherwise default to LIME under moderate or low privacy. For \textit{utility} (high $\alpha$), LIME is the sweet spot across all but the strictest privacy settings.

Based on the analysis of Figure 1, \textit{where we compare scores of each of the four explainers}, for applications requiring both strong privacy and explanation quality, LIME or SHAP combined with DPBart-1500 or DPPrompt-165 is advisable. In the case of SST2, aggressive privacy strategies such as TEM1 and TEM2 should be avoided, and moderate-DP setups using LIME are preferable. For AG News, LIME in combination with DPPrompt-165 or TEM3 performs robustly across all $\alpha$ values. Trustpilot demonstrates broad robustness and is a suitable candidate for real-world deployment in privacy-sensitive NLP scenarios.

When averaging \textit{across the four explainers}, our results show that using DPBart-1500 or DPPrompt-165 is recommended to achieve the best trade-off between privacy, utility, and explanation quality. TEM1 and TEM2 should be avoided when explanation faithfulness is a priority, especially for sensitive datasets such as SST2. For utility-focused applications (high $\alpha$), AG News paired with moderate-DP settings performs reliably well. In general, higher $\varepsilon$ values are preferable when maintaining interpretability is critical.

Based on the \textit{model size effect (base vs large model) analysis}, for privacy-preserving NLP tasks that \textit{require both high utility and faithful explanations}, \textit{base models} are consistently more reliable and effective than \textit{large models}. This is especially true when utility is prioritized (high $\alpha$) or for sensitive datasets such as \textit{SST2}.

These recommendations based on our experiments can be generalized into the following set of guidelines that are important to consider in the research or practice of \textit{explainable privacy} with natural language data:

\begin{enumerate}
    \itemsep 0em
    \item \textbf{Decide on the importance of privacy vs. utility}: an important starting point is the setting of $\alpha$, as this may be considerably different across various use cases.
    \item \textbf{Consider the nature of downstream task}: our results indicate that dataset (and task) are important factors in the juxtaposition of privacy and explainability. This becomes especially important in the following point.
    \item \textbf{Choose your privatization method wisely}: our initial findings suggest that for more complex tasks (e.g., multi-class classification), non-generative methods such as \textsc{TEM} may be more suitable. However, for more \say{colloquial} datasets, such as those stemming from user reviews, DP methods based on generative models may be more suitable to preserving both utility and explainability. While this guideline requires further validation, we emphasize the important interplay between choice of privatization method and nature of downstream task.
    \item \textbf{Choose the smallest pretrained model acceptable for the given use case}: we learn that across our results, composite scores decrease as model size increases, showing how smaller models, if acceptable for a given use case, may be preferable in finding a balance between privacy, utility, and explainability.
    \item \textbf{Measure on a variety of explainers}: we find that despite individual difference between explainers across all tested configurations, using averages and composite scores lead to clear emergent trends and interpretable differences between privacy levels ($\varepsilon$ values) and datasets/tasks. We therefore recommend the usage of multiple explainers, tied together by a composite score, for a robust overview of performance differences between setups.
\end{enumerate}

\section{Conclusion}
We conduct an investigation at the intersection of privacy and explainability, guided by the overarching methods of differentially private text rewriting and post-hoc explainability. In a series of experiments, we quantify the privacy-explainability trade-off, which lends interesting insights regarding the potential synergies between the two important topics. We are the first to conduct such an investigation in the context of natural language data, providing the foundations for further explorations into this interdisciplinary topic.

\paragraph{Future work.}
We envision a number of paths for future work based on our investigation, namely:
\begin{itemize}
    \itemsep 0em
    \item Further investigating \say{sweet spots}, particularly via the integration of more robust proxies for privacy (i.e., beyond $\varepsilon$ values). This could include for example the inclusion of membership inference testing, as performed by \citet{shokri_privacy_explanations_2021}.
    \item Extending our findings with experiments on additional datasets, Explainability and DP methods, and $\varepsilon$ ranges.
    \item Investigating the trade-offs with respect to non post-hoc explainability methods, as well as considering more recent frameworks for measuring faithfulness of explanations, such as that proposed by  \citet{Zheng_ffidelity_2025}.
    \item Including human evaluation (i.e, perceptions) into the calculation of composite score, namely to improve this score beyond automatic metrics.
\end{itemize}

\paragraph{Limitations.}
Our study has several limitations. It relies solely on quantitative evaluation, without qualitative or human assessments. We focus narrowly on post-hoc explainability and DP text rewriting methods, which do not fully capture either field. Our selection of DP methods overlooks nuanced differences in mechanism design and their varying theoretical guarantees. These constraints highlight the need for broader studies to deepen understanding of the explainability–privacy intersection in NLP.

\section*{Acknowledgments}
We thank the anonymous reviewers for their constructive feedback. This research has been supported by the German Federal Ministry of Education and Research (BMBF) grant 01IS23069 Software Campus 3.0 (TU München).

\bibliography{aaai25}

\end{document}